\documentclass[11pt,a4paper]{article}
\usepackage[hyperref]{acl2019}
\usepackage{times}
\usepackage{latexsym}
\usepackage{url}

\usepackage{multirow}
\usepackage{amsmath}
\usepackage{capt-of}
\usepackage{tabularx}
\usepackage[caption=false]{subfig}
\usepackage{epsfig}
\usepackage{amssymb}
\usepackage{amsfonts}
\usepackage{booktabs}
\usepackage{scalerel}
\usepackage[inline]{enumitem}
\usepackage{listings}
\usepackage{varwidth}
\usepackage[export]{adjustbox}
\usepackage{tikz}
\usetikzlibrary{tikzmark}
\usepackage{todonotes}
\usepackage{cleveref}

\usepackage{stmaryrd}
\usepackage{bbm}

\usepackage{algorithm}
\usepackage[noend]{algpseudocode}

\definecolor{deepblue}{rgb}{0,0,0.5}
\definecolor{officeblue}{RGB}{0,102,204}
\definecolor{deepred}{rgb}{0.6,0,0}
\definecolor{deepgreen}{rgb}{0,0.5,0}
\definecolor{mybrickred}{RGB}{182,50,28}

\definecolor{fillcolor}{RGB}{216,217,252}

\algnewcommand\algorithmicrequireb{{\hspace{0.85cm}}}
\algnewcommand\INPTDESCB{\item[\algorithmicrequireb]}

\algnewcommand\algorithmicfuncdesc{\textbf{Function:}}
\algnewcommand\FUNCDESC{\item[\algorithmicfuncdesc]}
\algnewcommand\algorithmicfuncdescb{{\hspace{1.48cm}}}
\algnewcommand\FUNCDESCB{\item[\algorithmicfuncdescb]}
\algnewcommand{\algorithmicgoto}{\textbf{goto}}
\algnewcommand{\Goto}[1]{\algorithmicgoto~\ref{#1}}

\usepackage{amsmath,amsfonts,bm}

\def\eqref#1{equation~\ref{#1}}

\def\1{\bm{1}}

\DeclareMathAlphabet{\mathsfit}{\encodingdefault}{\sfdefault}{m}{sl}
\SetMathAlphabet{\mathsfit}{bold}{\encodingdefault}{\sfdefault}{bx}{n}

\newcommand{\Ls}{\mathcal{L}}

\newcommand\mbase{$_{\textsc{base}}$}
\newcommand\mlarge{$_{\textsc{large}}$}
\newcommand\our{\textsc{MonoX}}
\newcommand\ourkd{\textsc{MonoX-Kd}}
\newcommand\ourpl{\textsc{MonoX-Pl}}

\aclfinalcopy 

\title{Can Monolingual Pretrained Models Help Cross-Lingual Classification?}

\author{Zewen Chi{$^\dag$}\thanks{\ \  Contribution during internship at Microsoft Research.},~~Li Dong\textsuperscript{$\ddagger$},~~Furu Wei\textsuperscript{$\ddagger$},~~Xian-Ling Mao\textsuperscript{$\dag$},~~Heyan Huang\textsuperscript{$\dag$} \\
\textsuperscript{$\dag$}Beijing Institute of Technology \\
\textsuperscript{$\ddagger$}Microsoft Research\\
\texttt{\{czw,maoxl,hhy63\}@bit.edu.cn}\\
\texttt{\{lidong1,fuwei\}@microsoft.com}}

\date{}

\begin{document}
\maketitle

\begin{abstract}
Multilingual pretrained language models (such as multilingual BERT) have achieved impressive results for cross-lingual transfer.
However, due to the constant model capacity, multilingual pre-training usually lags behind the monolingual competitors.
In this work, we present two approaches to improve zero-shot cross-lingual classification, by transferring the knowledge from monolingual pretrained models to multilingual ones.
Experimental results on two cross-lingual classification benchmarks show that our methods outperform vanilla multilingual fine-tuning.
\end{abstract}

\section{Introduction}

Supervised text classification heavily relies on manually annotated training data, while the data are usually only available in rich-resource languages, such as English. 
It requires great effort to make the resources available in other languages.
Various methods have been proposed to build cross-lingual classification models by exploiting machine translation systems~\cite{cldc,chen2018adversarial,xnli}, and learning multilingual embeddings~\cite{xnli,yu2018multilingual,clse-artetxet,eisenschlos2019multifit}.

Recently, multilingual pretrained language models have shown surprising cross-lingual effectiveness on a wide range of downstream tasks~\cite{bert,xlm,xnlg,xlmr}.
Even without using any parallel corpora, the pretrained models can still perform zero-shot cross-lingual classification~\cite{pires2019multilingual,wu2019beto,keung2019adversarial}.
That is, these models can be fine-tuned in a source language, and then directly evaluated in other target languages.
Despite the effectiveness of cross-lingual transfer, the multilingual pretrained language models have their own drawbacks.
Due to the constant number of model parameters, the model capacity of the rich-resource languages decreases if we adds languages for pre-training.
The curse of multilinguality results in that the multilingual models usually perform worse than their monolingual competitors on downstream tasks~\cite{nmt:wild19,xlmr}.
The observations motivate us to leverage monolingual pretrained models to improve multilingual models for cross-lingual classification.


In this paper, we propose a multilingual fine-tuning method (\our{}) based on the teacher-student framework, where a multilingual student model learns end task skills from a monolingual teacher.
Intuitively, monolingual pretrained models are used to provide supervision of downstream tasks, while multilingual models are employed for knowledge transfer across languages.
We conduct experiments on two widely used cross-lingual classification datasets,
where our methods outperform baseline models on zero-shot cross-lingual classification.
Moreover, we show that the monolingual teacher model can help the student multilingual model for both the source language and target languages, even though the student model is only trained in the source language.

\section{Background: Multilingual Fine-Tuning}

\begin{figure}[t]
\begin{center} 
\includegraphics[width=1.0\linewidth]{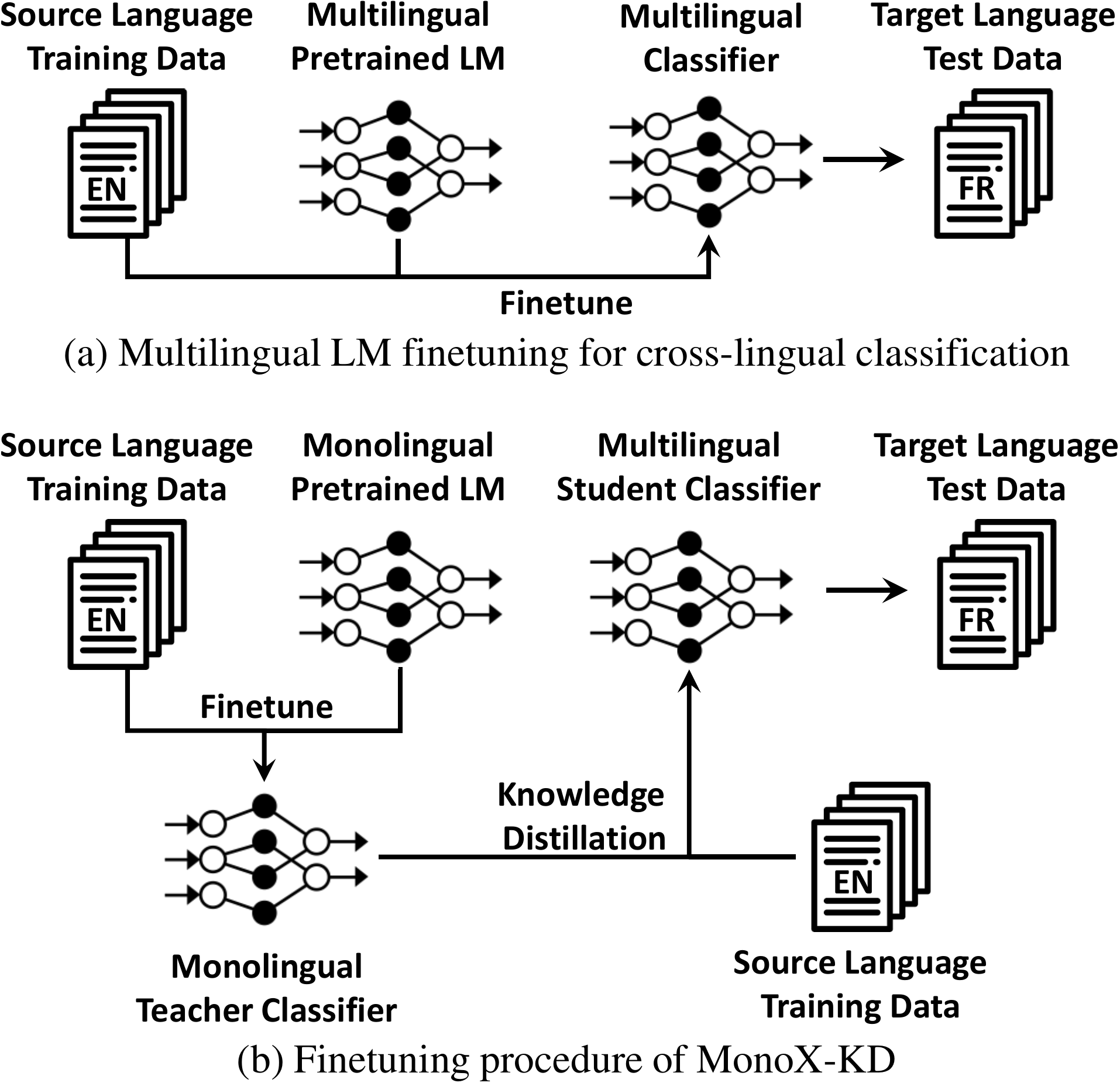}
\caption{Illustration of multilingual LM fine-tuning. (a) The original multilingual LM fine-tuning procedure for cross-lingual classification. (b) The fine-tuning procedure of our proposed \our{} via knowledge distillation (\ourkd{}). Notice that \our{} does not use any target language data during fine-tuning.
}
\label{fig:monox}
\end{center} 
\end{figure}

We use multilingual BERT~\cite{bert} for multilingual pretrained language models.
The pretrained model uses the BERT-style Transformer~\cite{transformer} architecture, and follows the similar fine-tuning procedure as BERT for text classification, which is illustrated in Figure~\ref{fig:monox}(a).
To be specific, the first input token of the models is always a special classification token \texttt{[CLS]}.
During fine-tuning, the final hidden state of the special token is used as the sentence representation.
In order to output predictions, an additional softmax classifier is built on top of the sentence representation.
Denoting $\mathcal{D}$ as the training data in the source language, the pretrained models are fine-tuned with standard cross-entropy loss:
\begin{align}
\Ls_{\textnormal{CE}}(\theta ; \mathcal{D}) = -\sum_{ (x,y) \in \mathcal{D}} \log p(y|x;\theta) \nonumber
\end{align}
where $\theta$ represents model parameters.
Then the model is directly evaluated on other languages for cross-lingual classification.

\section{Methods}

As shown in Figure~\ref{fig:monox}(b), we first fine-tune the monolingual pretrained model in the source language.
Then we transfer task knowledge to the multilingual pretrained model by soft (Section~\ref{sec:method:kd}) or hard (Section~\ref{sec:method:pl}) labels.
We describe two variants of our proposed method (\our{}) as follows.

\subsection{Knowledge Distillation}
\label{sec:method:kd}

In order to transfer task-specific knowledge from monolingual model to multilingual model, we propose to use knowledge distillation~\cite{knowledge:distill} under our \our{} framework, where a student model $s$ is trained with soft labels generated by a better-learned teacher model $t$.
The loss function of the student model is:
\begin{align}
&\Ls_{\textnormal{KD}}(\theta_s;\mathcal{D},\theta_t) = \nonumber\\
&-\sum_{ (x,y) \in \mathcal{D}} \sum_{k=1}^{K} q(y=k|x;\theta_t) \log p(y=k|x;\theta_s) \nonumber
\end{align}
where $p(\cdot)$ and $q(\cdot)$ represent the probability distribution over $K$ categories, predicted by the student $s$ and the teacher $t$, respectively.
Notice that only the student model parameters $\theta_s$ are updated during knowledge distillation.
As shown in Figure~\ref{fig:monox}(b), we first use the fine-tuned monolingual pretrained model as a teacher, which is learned by minimizing $\Ls_{\textnormal{CE}}(\theta_t;\mathcal{D})$. 
Then we perform knowledge distillation for the student model with $\Ls_{\textnormal{KD}}(\theta_s;\mathcal{D}_C,\theta_t)$ as the loss function, where $\mathcal{D}_C$ is the concatenation of training dataset and the unlabeled dataset in the source language.
We denote this implementation as \ourkd{}.

\subsection{Pseudo-Label}
\label{sec:method:pl}

In addition to knowledge distillation, we also consider implementing \our{} by training the student multilingual model with pseudo-label~\cite{pseudo:label}.
Specifically, after fine-tuning the monolingual pretrained model on the training data as teacher, we apply the teacher model on the unlabeled data in the source language to generate pseudo labels.
Next, we filter the pseudo labels by a prediction confidence threshold, and only keep the examples with higher confidence scores.
Notice that the pseudo training data are assigned with hard labels.
Finally, we concatenate the original training data and the pseudo data as the final training set for the student model.
We denote this implementation as \ourpl{}.

\section{Experiments}

\subsection{Experimental Setup}

In the following experiments, we consider the zero-shot cross-lingual setting, where models are trained with English data and directly evaluated on all target languages.

\paragraph{Datasets}
We conduct experiments on two widely used datasets for cross-lingual evaluation:
(1) Cross-Lingual Sentiment (\textsc{CLS}) dataset~\cite{cls}, containing Amazon reviews in three domains and four languages;
(2) Cross-Lingual NLI (\textsc{XNLI}) dataset~\cite{xnli}, containing development and test sets in 15 languages and a training set in English for the natural language inference task.

\paragraph{Pretrained Language Models}
We use multilingual BERT\mbase{}\footnote{\url{https://github.com/google-research/bert/blob/master/multilingual.md}} for cross-lingual transfer.
For monolingual pretrained language model, the English-version RoBERTa\mlarge{}\footnote{\url{https://github.com/pytorch/fairseq/tree/master/examples/roberta}} is employed.
All the pretrained models used in our experiments are cased models.

\paragraph{Baselines}
We compare our methods (\ourkd{}, and \ourpl{}) with the following models:
\begin{itemize}
\item \textsc{mBert}: directly fine-tuning the multilingual BERT\mbase{} with English training data.
\item \textsc{mBert-St}: fine-tuning the multilingual BERT\mbase{} by self-training, i.e., alternately fine-tuning mBERT and updating the training data by labeling English unlabeled examples.
\end{itemize}

\subsection{Configuration}
For the \textsc{CLS} dataset, we randomly select 20\% examples from training data as the development set and use the remaining examples as the training set.
For \textsc{XNLI}, we randomly sample 20\% examples from training data as the training set, and regard the other examples as the unlabeled set.
We use the vocabularies provided by the pretrained models, which are extracted by Byte-Pair Encoding \cite{bpe}. The input sentences are truncated to 256 tokens.
For both datasets, we use Adam optimizer with a learning rate of $5\times10^{-6}$, and a batch size of 8. We train models with epoch size of 200 and 2,500 steps for \textsc{CLS} and \textsc{XNLI}, respectively. For \ourkd{}, the softmax temperature of knowledge distillation is set to $0.1$. For \ourpl{}, the confidence threshold is set to zero, which means all of the generated pseudo labels are used as training data.

\begin{table}[t]
\centering
\begin{tabular}{lccc}
\toprule
& Parameters & CLS & \textsc{XNLI} \\ \midrule
\multicolumn{3}{l}{\textit{Multilingual Pretrained Models}} & \\
\textsc{mBert} & 110M & 86.37 & 77.07\\ \midrule
\multicolumn{3}{l}{\textit{Monolingual Pretrained Models}} & \\
\textsc{Bert}\mbase{} & 110M & 90.10 & 80.46 \\
RoBERTa\mbase{} & 125M & 93.82 & 85.09\\
RoBERTa\mlarge{} & 355M & 95.77 & 89.24 \\
\bottomrule
\end{tabular}
\caption{Preliminary experiments results. Models are finetuned with English training data of \textsc{CLS} and \textsc{XNLI} under the configuration (see Section 3.2), and only evaluated in English. The results on \textsc{CLS} are averaged over three domains.}
\label{table:pre}
\end{table}

\begin{table*}[t]
\centering
\small
\renewcommand\tabcolsep{4.0pt}
\begin{tabular}{l|ccc|ccc|ccc|ccc|c}
\toprule
&  \multicolumn{3}{c|}{en} &  \multicolumn{3}{c|}{de} &  \multicolumn{3}{c|}{fr} &  \multicolumn{3}{c|}{ja} \\
&  Books &    DVD &  Music &  Books &    DVD &  Music &  Books &    DVD &  Music &  Books &    DVD &  Music &       avg \\ \midrule
\textsc{mBert}    &     87.75 &     86.60 &     84.75 &     79.55 &     75.90 &     77.05 &     81.45 &     80.35 &     80.35 &     75.15 &     76.90 &     75.90 &     80.14 \\
\textsc{mBert-St} &     88.20 &     85.50 &     88.00 &     79.65 &     76.70 &     80.00 &     84.85 &     83.25 &     80.55 &     74.60 &     75.80 &     76.90 &     81.17 \\ \midrule
\ourpl{}          &\textbf{94.00}&     \textbf{92.75} &     91.80 &     83.20 &     79.25 &     82.95 &     \textbf{86.00} &     84.95 &     \textbf{84.55} &     78.85 &     \textbf{80.00} &     79.35 &     84.80 \\
\ourkd{}          &     93.90 &     91.40 &     \textbf{92.25} &     \textbf{84.20} &     \textbf{81.50} &     \textbf{83.65} &     85.40 &     \textbf{85.90} &     83.95 &     \textbf{78.95} &     79.15 &     \textbf{80.30} &     \textbf{85.05} \\
\bottomrule
\end{tabular}
\caption{Evaluation results of zero-shot cross-lingual sentiment classification on the \textsc{CLS} dataset.}
\label{table:cls}
\end{table*}

\begin{table*}[t]
\centering
\small
\renewcommand\tabcolsep{4.0pt}
\begin{tabular}{lcccccccccccccccc}
\toprule
&   ar &   bg &   de &   el &   en &   es &   fr &   hi &   ru &   sw &   th &   tr &   ur &   vi &   zh &  avg\\ \midrule
\textsc{mBert}    & 61.2 & 67.4 & 65.8 & 61.6 & 77.1 & 70.7 & 68.6 & \textbf{53.4} & 67.0 & 50.6 & \textbf{44.6} & 56.3 & 57.8 & 43.6 & 67.8 & 60.9 \\
\textsc{mBert-St} & 60.9 & 67.6 & 65.4 & 61.0 & 77.6 & 70.4 & 68.9 & 53.1 & 65.9 & 50.6 & 41.8 & 55.2 & 56.8 & 43.6 & 67.9 & 60.5 \\ \midrule
\ourpl{}          & \textbf{63.5} & \textbf{70.1} & \textbf{69.8} & 61.7 & \textbf{80.9} & \textbf{74.1} & \textbf{72.1} & 52.5 & 68.4 & 51.2 & 42.3 & \textbf{57.9} & 58.0 & 44.0 & 70.2 & \textbf{62.5} \\
\ourkd{}          & 62.2 & 69.3 & 69.3 & \textbf{62.1} & 79.6 & 72.9 & 72.0 & 52.8 & \textbf{68.6} & \textbf{52.3} & 41.7 & \textbf{57.9} & \textbf{58.5} & \textbf{45.9} & \textbf{70.8} & 62.4 \\
\bottomrule
\end{tabular}
\caption{Evaluation results of zero-shot cross-lingual NLI on the \textsc{XNLI} dataset. Note that 20\% of the original training data are used as training set, and the other 80\% are used as unlabeled set.}
\label{table:xnli}
\end{table*}

\subsection{Results and Discussion}


\paragraph{Preliminary Experiments}
To see how much monolingual pretrained models is better than multilingual pretrained models, we finetune several different pretrained language models on the two datasets under the aforementioned configuration, and only evaluate them in English. As shown in Table~\ref{table:pre}, the gap between multilingual and monolingual pretrained models is large, even when using the same size of parameters. It is not hard to explain because \textsc{mBert} is trained in 104 languages, 
where different languages tend to confuse each other.

\paragraph{Sentiment Classification} 
We evaluate our method on the zero-shot cross-lingual sentiment classification task. The goal of sentiment classification is to classify input sentences to positive or negative sentiments.
In Table~\ref{table:cls} we compare the results of our methods with baselines on \textsc{CLS}. It can be observed that our \our{} method outperforms baselines in all evaluated languages and domains, providing 4.91\% improvement of averaged accuracy to the original multilingual BERT fine-tuning method. 
Notice that \textsc{mBert-St} is trained under the same condition with our method, i.e., using the same labeled and unlabeled data as ours. 
However, we only observe a slight improvement over \textsc{mBert},
which demonstrates that the performance improvement of \our{} mainly benefits from its end task knowledge transfer rather than the unlabeled data.

\paragraph{Natural Language Inference}
We also evaluate our method on the zero-shot cross-lingual NLI task, which is more challenging than sentiment classification. The goal of NLI is to identify the relationship of a pair of input sentences, including a premise and a hypothesis with an \textit{entailment}, \textit{contradiction}, or \textit{neutral} relationship between them. As shown in Table~\ref{table:xnli}, we present the evaluation results on \textsc{XNLI}. Unsurprisingly, both \ourpl{} and \ourkd{} perform better than baseline methods,
showing that our method successfully helps the multilingual pretrained model gain end task knowledge from the monolingual pretrained model for cross-lingual classification.
It is also worth mentioning that the performance of \textsc{mBert-St} is similar to \textsc{mBert}. We believe the reason is that \textsc{XNLI} has more training data than \textsc{CLS}, which wakens the impact of self-training.

\paragraph{Effects of Training Data Size}

\begin{figure}[t]
\begin{center} 
\includegraphics[width=1.0\linewidth]{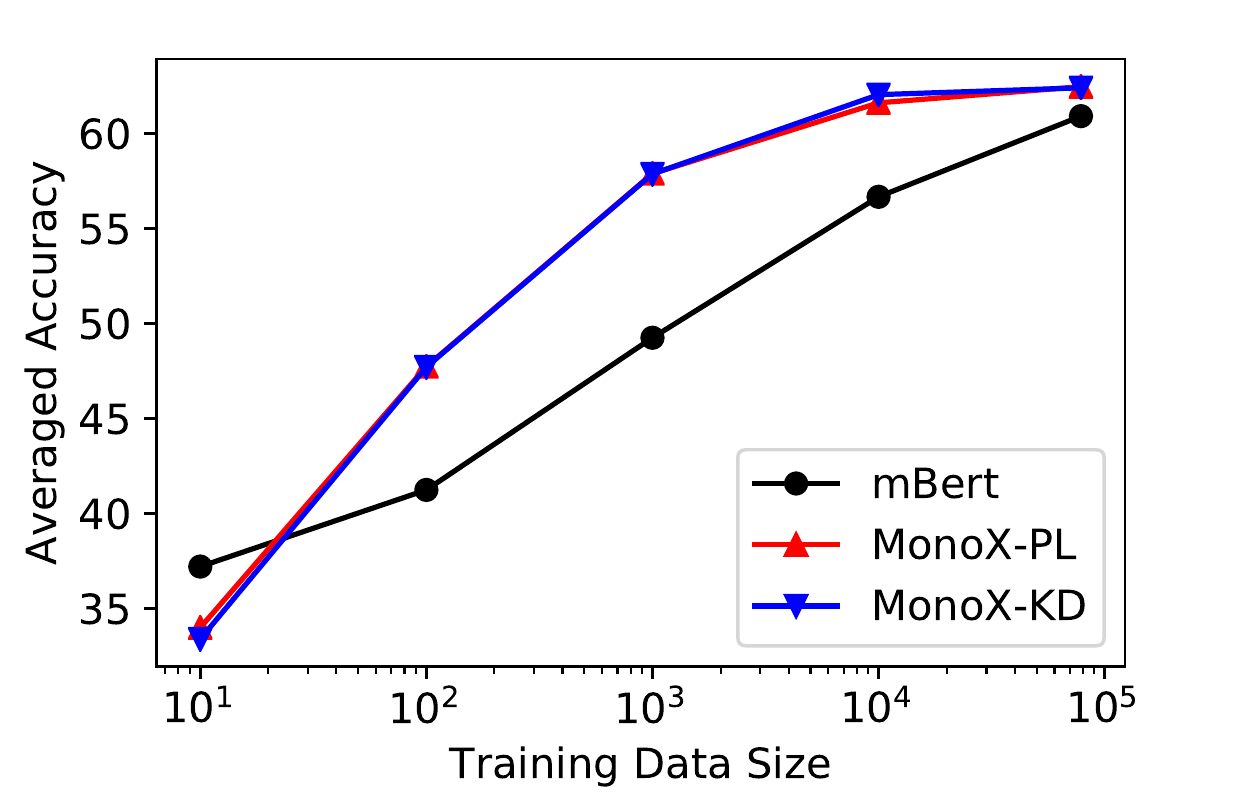}
\caption{Averaged accuracy scores on zero-shot \textsc{XNLI} with different training data sizes. (20\% and 80\% of the training data are regraded training and unlabeled set.)
} 
\label{fig:data}
\end{center} 
\end{figure}

We conduct a study on how much multilingual pretrained model can learn from monolingual pretrained model for different training data size. We cut the training data to 10, 100, 1K, 10K and 78K (full training data in our setting) examples, and keep other hyper-parameters fixed. In Figure~\ref{fig:data}, we show the averaged accuracy scores for zero-shot \textsc{XNLI} with different training data sizes. We observe that \our{} outperforms \textsc{mBert} on all data sizes except the 10-example setting. When the training data is relatively small ($\leq 10^4$), our method shows a great improvement.

\paragraph{Effects of Distillation Temperature}

\begin{figure}[t]
\begin{center} 
\includegraphics[width=1.0\linewidth]{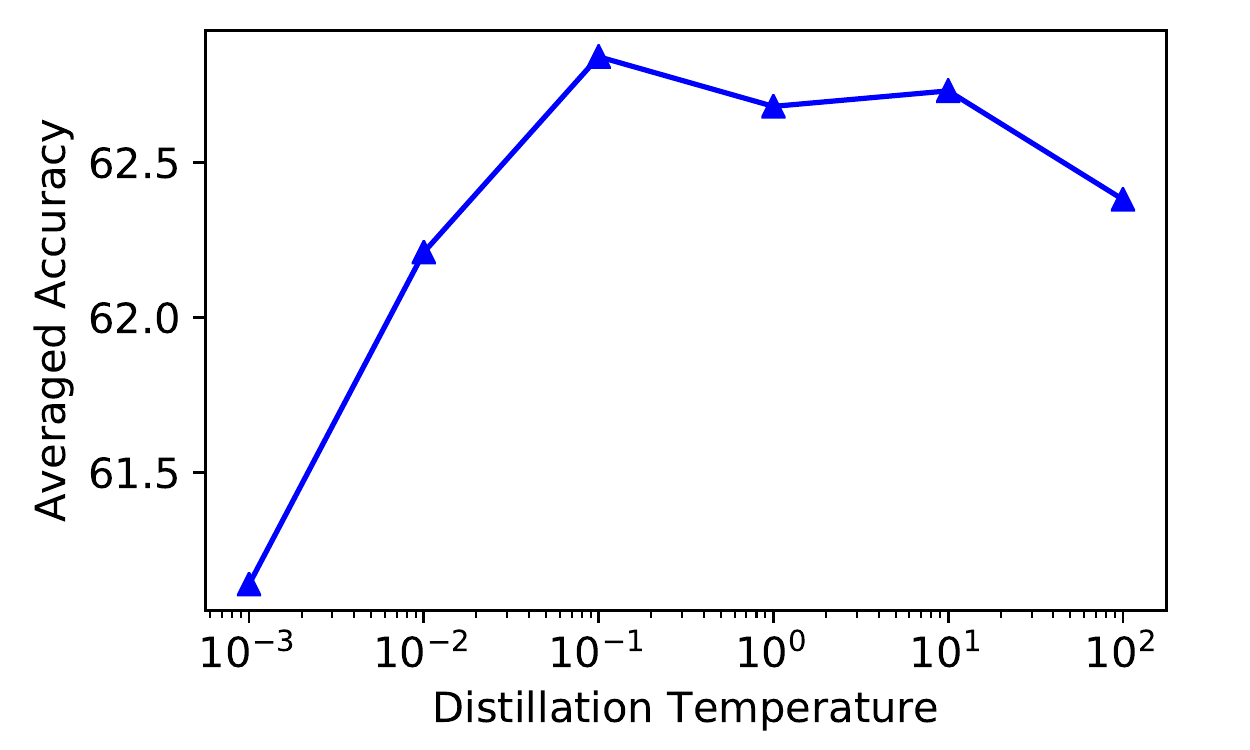}
\caption{Averaged accuracy scores on the development set for zero-shot \textsc{XNLI} with different softmax temperatures of \ourkd{}.
} 
\label{fig:temp}
\end{center} 
\end{figure}

Figure~\ref{fig:temp} presents \textsc{XNLI} averaged accuracy scores of \ourkd{} with different softmax temperatures in knowledge distillation. Even though the temperature varies from $10^{-3}$ to $10^2$, all of the results are higher than baseline scores, which indicates \ourkd{} is nonsensitive to the temperature. When the temperature is set to $10^{-1}$, we observe the best results on the development set. Therefore we set temperature as $0.1$ in other experiments.

\section{Conclusion}

In this work, we investigated whether a monolingual pretrained model can help cross-lingual classification.
Our results have shown that, with a RoBERTa model pretrained in English, we can boost the classification performance of a pretrained multilingual BERT in other languages.
For future work, we will explore whether monolingual pretrained models can help other cross-lingual NLP tasks, such as natural language generation.

\bibliography{xft}

\begin{thebibliography}{19}
\expandafter\ifx\csname natexlab\endcsname\relax\def\natexlab#1{#1}\fi

\bibitem[{Arivazhagan et~al.(2019)Arivazhagan, Bapna, Firat, Lepikhin, Johnson,
  Krikun, Chen, Cao, Foster, Cherry, Macherey, Chen, and Wu}]{nmt:wild19}
Naveen Arivazhagan, Ankur Bapna, Orhan Firat, Dmitry Lepikhin, M.~Gatu Johnson,
  Maxim Krikun, Mia~Xu Chen, Yuan Cao, George Foster, Colin Cherry, Wolfgang
  Macherey, Zhifeng Chen, and Yonghui Wu. 2019.
\newblock Massively multilingual neural machine translation in the wild:
  Findings and challenges.
\newblock \emph{ArXiv}, abs/1907.05019.

\bibitem[{Artetxe and Schwenk(2019)}]{clse-artetxet}
Mikel Artetxe and Holger Schwenk. 2019.
\newblock Massively multilingual sentence embeddings for zero-shot
  cross-lingual transfer and beyond.
\newblock \emph{Transactions of the Association for Computational Linguistics},
  7:597--610.

\bibitem[{Chen et~al.(2018)Chen, Sun, Athiwaratkun, Cardie, and
  Weinberger}]{chen2018adversarial}
Xilun Chen, Yu~Sun, Ben Athiwaratkun, Claire Cardie, and Kilian Weinberger.
  2018.
\newblock Adversarial deep averaging networks for cross-lingual sentiment
  classification.
\newblock \emph{Transactions of the Association for Computational Linguistics},
  6:557--570.

\bibitem[{Chi et~al.(2019)Chi, Dong, Wei, Wang, Mao, and Huang}]{xnlg}
Zewen Chi, Li~Dong, Furu Wei, Wenhui Wang, Xian-Ling Mao, and Heyan Huang.
  2019.
\newblock Cross-lingual natural language generation via pre-training.
\newblock \emph{arXiv preprint arXiv:1909.10481}.

\bibitem[{Conneau et~al.(2019)Conneau, Khandelwal, Goyal, Chaudhary, Wenzek,
  Guzmán, Grave, Ott, Zettlemoyer, and Stoyanov}]{xlmr}
Alexis Conneau, Kartikay Khandelwal, Naman Goyal, Vishrav Chaudhary, Guillaume
  Wenzek, Francisco Guzmán, Edouard Grave, Myle Ott, Luke Zettlemoyer, and
  Veselin Stoyanov. 2019.
\newblock Unsupervised cross-lingual representation learning at scale.
\newblock \emph{ArXiv}, abs/1911.02116.

\bibitem[{Conneau et~al.(2018)Conneau, Lample, Rinott, Williams, Bowman,
  Schwenk, and Stoyanov}]{xnli}
Alexis Conneau, Guillaume Lample, Ruty Rinott, Adina Williams, Samuel~R Bowman,
  Holger Schwenk, and Veselin Stoyanov. 2018.
\newblock Xnli: Evaluating cross-lingual sentence representations.
\newblock \emph{arXiv preprint arXiv:1809.05053}.

\bibitem[{Devlin et~al.(2018)Devlin, Chang, Lee, and Toutanova}]{bert}
Jacob Devlin, Ming{-}Wei Chang, Kenton Lee, and Kristina Toutanova. 2018.
\newblock {BERT:} pre-training of deep bidirectional transformers for language
  understanding.
\newblock \emph{CoRR}, abs/1810.04805.

\bibitem[{Eisenschlos et~al.(2019)Eisenschlos, Ruder, Czapla, Kardas, Gugger,
  and Howard}]{eisenschlos2019multifit}
Julian Eisenschlos, Sebastian Ruder, Piotr Czapla, Marcin Kardas, Sylvain
  Gugger, and Jeremy Howard. 2019.
\newblock Multifit: Efficient multi-lingual language model fine-tuning.
\newblock \emph{arXiv preprint arXiv:1909.04761}.

\bibitem[{Hinton et~al.(2015)Hinton, Vinyals, and Dean}]{knowledge:distill}
Geoffrey~E. Hinton, Oriol Vinyals, and Jeffrey Dean. 2015.
\newblock Distilling the knowledge in a neural network.
\newblock \emph{ArXiv}, abs/1503.02531.

\bibitem[{Keung et~al.(2019)Keung, Lu, and Bhardwaj}]{keung2019adversarial}
Phillip Keung, Yichao Lu, and Vikas Bhardwaj. 2019.
\newblock Adversarial learning with contextual embeddings for zero-resource
  cross-lingual classification and ner.
\newblock \emph{arXiv preprint arXiv:1909.00153}.

\bibitem[{Lample and Conneau(2019)}]{xlm}
Guillaume Lample and Alexis Conneau. 2019.
\newblock Cross-lingual language model pretraining.
\newblock \emph{arXiv preprint arXiv:1901.07291}.

\bibitem[{Lee(2013)}]{pseudo:label}
Dong-Hyun Lee. 2013.
\newblock Pseudo-label : The simple and efficient semi-supervised learning
  method for deep neural networks.
\newblock \emph{ICML 2013 Workshop : Challenges in Representation Learning}.

\bibitem[{Pires et~al.(2019)Pires, Schlinger, and
  Garrette}]{pires2019multilingual}
Telmo Pires, Eva Schlinger, and Dan Garrette. 2019.
\newblock How multilingual is multilingual bert?
\newblock \emph{arXiv preprint arXiv:1906.01502}.

\bibitem[{Prettenhofer and Stein(2010)}]{cls}
Peter Prettenhofer and Benno Stein. 2010.
\newblock Cross-language text classification using structural correspondence
  learning.
\newblock In \emph{Proceedings of the 48th annual meeting of the association
  for computational linguistics}, pages 1118--1127.

\bibitem[{Sennrich et~al.(2015)Sennrich, Haddow, and Birch}]{bpe}
Rico Sennrich, Barry Haddow, and Alexandra Birch. 2015.
\newblock Neural machine translation of rare words with subword units.
\newblock \emph{arXiv preprint arXiv:1508.07909}.

\bibitem[{Vaswani et~al.(2017)Vaswani, Shazeer, Parmar, Uszkoreit, Jones,
  Gomez, Kaiser, and Polosukhin}]{transformer}
Ashish Vaswani, Noam Shazeer, Niki Parmar, Jakob Uszkoreit, Llion Jones,
  Aidan~N Gomez, {\L}ukasz Kaiser, and Illia Polosukhin. 2017.
\newblock \href
  {http://papers.nips.cc/paper/7181-attention-is-all-you-need.pdf} {Attention
  is all you need}.
\newblock In \emph{Advances in Neural Information Processing Systems 30}, pages
  5998--6008. Curran Associates, Inc.

\bibitem[{Wu and Dredze(2019)}]{wu2019beto}
Shijie Wu and Mark Dredze. 2019.
\newblock Beto, bentz, becas: The surprising cross-lingual effectiveness of
  bert.
\newblock \emph{arXiv preprint arXiv:1904.09077}.

\bibitem[{Xu and Yang(2017)}]{cldc}
Ruochen Xu and Yiming Yang. 2017.
\newblock Cross-lingual distillation for text classification.
\newblock \emph{arXiv preprint arXiv:1705.02073}.

\bibitem[{Yu et~al.(2018)Yu, Li, and Oguz}]{yu2018multilingual}
Katherine Yu, Haoran Li, and Barlas Oguz. 2018.
\newblock Multilingual seq2seq training with similarity loss for cross-lingual
  document classification.
\newblock In \emph{Proceedings of The Third Workshop on Representation Learning
  for NLP}, pages 175--179.

\end{thebibliography}
\bibliographystyle{acl_natbib}

\end{document}